\documentclass[nohyperref]{article}

\usepackage{microtype}
\usepackage{graphicx}
\usepackage{booktabs} 
\usepackage{multirow}
\usepackage{caption}
\usepackage{subcaption}

\usepackage{hyperref}

\usepackage[accepted]{icml2022}

\usepackage{amsmath}
\usepackage{amssymb}
\usepackage{mathtools}
\usepackage{amsthm}

\usepackage[capitalize,noabbrev]{cleveref}

\theoremstyle{plain}

\theoremstyle{definition}

\theoremstyle{remark}

\usepackage[textsize=tiny]{todonotes}
\usepackage[acronym,toc,shortcuts]{glossaries}
\newacronym{DP}{DP}{Differential Privacy}
\newacronym{DPSGD}{DP-SGD}{Differentially Private Stochastic Gradient Descent}
\newacronym{BN}{BN}{Batch Normalisation}
\newacronym{IN}{IN}{Instance Normalisation}
\newacronym{LN}{LN}{Layer Normalisation}
\newacronym{GN}{GN}{Group Normalisation}
\newacronym{SOTA}{SOTA}{State-of-the-art}
\hypersetup{final}
\icmltitlerunning{SmoothNets: Optimizing CNN architecture design for differentially private deep learning}

\begin{document}

\twocolumn[
\icmltitle{SmoothNets: Optimizing CNN architecture design for differentially private deep learning}



\icmlsetsymbol{equal}{*}

\begin{icmlauthorlist}
\icmlauthor{Nicolas W. Remerscheid}{equal,tum}
\icmlauthor{Alexander Ziller}{equal,tum,rad}
\icmlauthor{Daniel Rueckert}{tum,icl}
\icmlauthor{Georgios Kaissis}{tum,rad,icl}
\end{icmlauthorlist}

\icmlaffiliation{tum}{Chair for AI in Medicine, Technical University of Munich, Germany}
\icmlaffiliation{rad}{Department of Radiology, Technical University of Munich, Germany}
\icmlaffiliation{icl}{Department of Computing, Imperial College London, United Kingdom}

\icmlcorrespondingauthor{Georgios Kaissis}{g.kaissis@tum.de}

\icmlkeywords{Differential Privacy, Deep Learning, Architecture Design}

\vskip 0.3in
]



\printAffiliationsAndNotice{\icmlEqualContribution} 

\begin{abstract}
The arguably most widely employed algorithm to train deep neural networks with Differential Privacy is \gls*{DPSGD}, which requires clipping and noising of per-sample gradients. This introduces a reduction in model utility compared to non-private training. Empirically, it can be observed that this accuracy degradation is strongly dependent on the model architecture. We investigated this phenomenon and, by combining components which exhibit good individual performance, distilled a new model architecture termed SmoothNet, which is characterised by increased robustness to the challenges of DP-SGD training. Experimentally, we benchmark SmoothNet against standard architectures on two benchmark datasets and observe that our architecture outperforms others, reaching an accuracy of 73.5\% on CIFAR-10 at $\varepsilon=7.0$ and 69.2\% at $\varepsilon=7.0$ on ImageNette, a state-of-the-art result compared to prior architectural modifications for \gls*{DP}.
\end{abstract}

\section{Introduction}
\gls{DPSGD}, introduced by \citeauthor{abadi2016deep}, enables training deep learning models with Differential Privacy (DP). It offers a guarantee to individuals whose data is used to train the model that the information which can be learned about them is limited. Often, \acs{DPSGD} comes at the cost of impaired model utility. This trade-off is especially undesirable in critical domains as medicine, where the unfortunate requirement of choosing between strong privacy and high-quality predictions is introduced. Prior work suggests that various model architectures provide very different robustness to such a utility drop \cite{papernot2019making, morsbach2021architecture, ziller2021differentially}. These findings lead to the hypothesis that the presence of \textit{specific architectural components} in deep learning architectures could influence their overall robustness against the specific challenges of \acs{DPSGD}, namely a geometric bias induced by per-sample gradient clipping \citep{chen2020understanding} and an overall noisy optimisation trajectory \citep{dormann2021not}.     
To investigate this conjecture, we systematically evaluated specific components of standard architectures and their influence on \acs{DP}-robustness, concretely residual connections \citep{he2016deep}, dense connections \citep{huang2017densely} and, inspired by \citet{tan2019efficientnet}, the depth and width of the architecture's convolutional layers. From the conclusions gained, we distilled a model architecture for computer vision tasks, specifically designed to withstand the challenges of \acs{DP}-Training, which we term SmoothNet.
Our key contributions are:
\begin{itemize}
    \item We systematically evaluated individual model components of popular deep learning architectures in regards to their influence on the performance of \acs{DPSGD} training.
    \item From this we distilled \textit{optimal} components and assembled a new model architecture, termed \textit{SmoothNet}, which yields SOTA results in terms of differentially privately trained models on benchmark datasets CIFAR-10 and ImageNette. 
\end{itemize}

\subsection{Prior Work}
\label{sec:prior_work}
Several lines of work have attempted to tackle the utility ramifications of deep network training with \acl{DP} guarantees. They can be largely categorised into two types of approaches: (1) \textbf{Architectural modifications}, where model architectures are adapted to be robust against the challenges in \acs{DPSGD}-based training and (2) \textbf{training methods}, which propose algorithmic modifications to the model training process. We provide an overview of these results categorised by approach in Table \ref{tab:other_works}. 
\begin{table*}[ht]
    \centering
    \begin{tabular}{@{}lllll@{}}
    \toprule
        Reference & Dataset & $\varepsilon$ & Accuracy & Type of work \\ \midrule
        \citealt{papernot2020tempered} & \multirow{4}{*}{CIFAR-10} & $7.53$ & $66.2\%$ & Architecture \\
        \citealt{klause2022differentially} & & $7.5$ & $71.7\%$ & Architecture \\
        \citealt{dormann2021not} & & $7.42$ & $70.1\%$ & Training method \\
        \citealt{de2022unlocking} & & $8.0$ & $81.4\%$ & Training method \\ \midrule
        \citealt{klause2022differentially} & ImageNette & $7.5$ & $64.8\%$ & Architecture\\
    \bottomrule
    \end{tabular}
    \caption{Results of other works on the same datasets utilised in our study. \textit{Type of work} refers to the categories specified in Section \ref{sec:prior_work}.}
    \label{tab:other_works}
\end{table*}
\paragraph{Architectural modifications}

\citeauthor{papernot2020tempered} proposed \textit{Tempered Sigmoids} as activation functions for \acs{DPSGD} training, which they applied in small feed-forward convolutional neural networks on the MNIST, FashionMNIST and CIFAR-10 datasets. \citeauthor{klause2022differentially} adapted the ResNet architecture \cite{he2016deep} and introduced an additional normalization layer after each residual block. Authors argue that the scale mismatch between the activations output by the convolutional and residual paths of the architecture inhibit convergence when trained with \acs{DP}, especially since --contrary to non-private training-- Batch Normalisation layers \cite{ioffe2015batch} are not available.

\paragraph{Training methods}
\citeauthor{dormann2021not} showed that the utility of \acs{DPSGD} training crucially depends on the batch size used during training \cite{dormann2021not}. They empirically demonstrated that higher batch sizes allow for better model utility at the same privacy budget levels, as the sub-sampling amplification guarantees of \acs{DPSGD} get outweighed by the stabilisation of the optimisation trajectory due to a decrease in gradient bias. This finding was underscored by the very recent work by \citeauthor{de2022unlocking}, which was released during the preparation of this manuscript. This work also showed the positive impact of weight standardization \cite{qiao2019micro}, parameter averaging \cite{polyak1992acceleration, tan2019efficientnet} and lastly \textit{augmentation multiplicity} \cite{fort2021drawing, hoffer2019augment, touvron2021training}, where each sample is augmented multiple times and an average gradient of all augmentations is treated as the gradient of this sample. These training modifications led to a new SOTA performance on the CIFAR-10 benchmark dataset \cite{krizhevsky2009learning} and (with pre-training on a proprietary dataset) on ImageNet \cite{deng2009imagenet}. 

\section{SmoothNets}
\begin{figure*}[ht]
    \centering
    \begin{subfigure}[b]{\textwidth}
    \centering
    \includegraphics[width=0.4\textwidth]{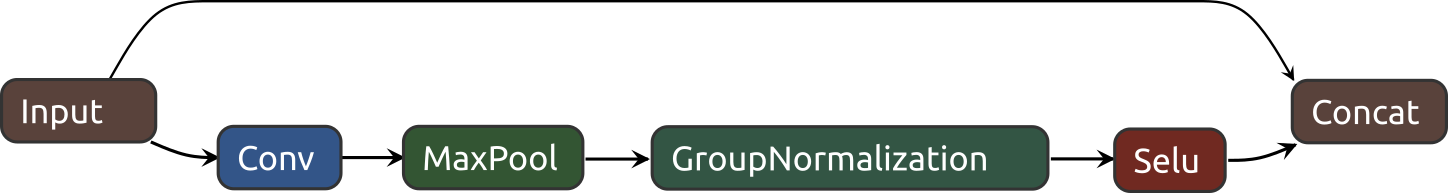}
    \caption{Main Building Block of SmoothNet. Similar to a DenseBlock, a residual connection concatenates the input and the processed features. However, we use wider convolutional layers, Group Normalization with 8 groups for DP compatibility, and a SELU activation function.}
    \label{fig:smoothblock}
    \end{subfigure}
    \begin{subfigure}[b]{\textwidth}
    \includegraphics[width=\textwidth]{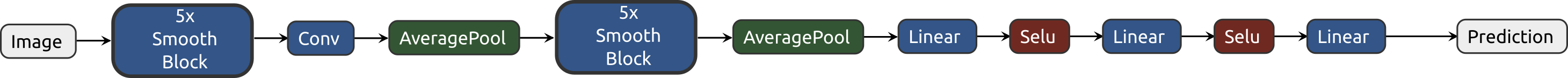}
    \caption{Architecture of the SmoothNet employed in our experiments consisting of SmoothBlocks, wide convolutional layers, average pooling layers, linear layers and SELU activation functions. $5\times$ is used as shorthand to denote 5 consecutive SmoothBlocks.}
    \label{fig:smoothnet}
    \end{subfigure}
    \caption{Overview of the building blocks for our SmoothNet architecture.}
    \label{fig:architecture}
\end{figure*}
So far, only few works have considered \textit{which concrete model design choices} provide robustness against utility drops for \acs{DPSGD} training. The specific challenges in this scenario originate from the requirements of ascertaining a bounded gradient norm and of adding noise to the gradients during training in order to realise DP guarantees. To shed more light on the correlation between model architecture and robustness/utility, we conducted preliminary experiments in which architectural design choices were isolated and systematically evaluated regarding their influence on the performance of DP training. 

\paragraph{Empirical study of popular model components}
In this work, we focused on determining the optimal ratio of \textit{depth} to \textit{width} (in terms of convolutional filters), a line of research initiated by \citet{tan2019efficientnet}. Moreover, we investigated \textit{residual connections} proposed by \citet{he2016deep}, and \textit{dense connections} proposed by \cite{huang2017densely}. In addition, we evaluated the effect of different \textit{normalization} layers, namely \textit{instance normalization} \cite{ulyanov2016instance}, \textit{group normalization} \cite{wu2018group} and \textit{layer normalization} \cite{ba2016layer} as optimal normalization components for \acs{DPSGD} training. Furthermore, we evaluated pooling layers (\textit{maximum pooling} and \textit{average pooling}) within the SmoothBlocks (compare Figure \ref{fig:smoothblock}) as well as activation functions (ReLU \cite{nair2010rectified}, SELU \cite{klambauer2017self}). In total, we evaluated over 100 different model architectures in more than 6000 experiments on CIFAR-10 based on (a) handcrafted models, intended to inspect the effect of \textit{isolated} model components and (b) on established (SOTA) models to guarantee \textit{representative} results. We use the SGD optimizer with a learning rate scheduler decaying the initial learning rate of $0.1$ by a factor of $0.7$ every epoch. Furthermore, we use a batch-size of 128, an early stopping patience of 15 epochs on the validation loss, a divergence threshold of 1000, and conducted a Bayesian optimization over the $L_2$ clipping norm in the range of $[0.1, 10.0]$ and over the number of epochs resulting in a binary decision of either 15 or 30 epochs. We note that we did not count the privacy budget of these hyperparameter searches against the overall allotted privacy budget.
We conducted 15 runs per sweep and considered the 10 best models ordered by validation accuracy (to suppress artifacts from poor initialization). For \acs{DPSGD} training we employed the Opacus \cite{Opacus} and deepee \cite{ziller2021medical} packages.

\paragraph{Observations}
From our experiments, we deduced two main lines of observations: (1) The width to depth ratio is strongly correlated with model performance. Concretely, \textbf{the optimal width-depth ratio is higher for private training compared to non-private training} and optimally $>1$, corresponding to a \textit{wide} model. (2) Both residual and dense connections are beneficial for robust models, with \textbf{dense connections being especially useful for training with DP-guarantees}. Furthermore, \textbf{Group Normalization proved to outperform other normalization layers}, which corroborates findings from prior works \cite{klause2022differentially, de2022unlocking}. The best-performing activation function in our experiments was SELU. Max Pooling as a SmoothBlock component led to superior results compared to other pooling functions. 

\paragraph{SmoothNet Architecture}
Based on the aforementioned conclusions, we synthesized the SmoothNet architecture. The main components are building blocks we term \textit{SmoothBlocks}. These are conceptually similar to DenseBlocks \cite{huang2017densely}, however with several modifications: (1) The $3\times 3$ convolutional layers rapidly increase in width such that, at the widest stage, \textit{more than 800} channels are present. (2) We use Group Normalisation layers with 8 groups, which replace Batch Normalisation in the original architecture. (3) SELU layers \cite{klambauer2017self} are used as activation functions. We restrict ourselves to a limited depth, of 10 SmoothBlocks. As in DenseNets \cite{huang2017densely}, we use average pooling between SmoothBlocks and to compress the extracted features to 2048 features, which are input to the classifying part. As a classifier block, we use three linear layers, separated by SELU activation functions. An overview of the SmoothBlock and the SmoothNet architecture can be found in Figure \ref{fig:architecture}. 

\section{Results} 
\begin{table*}[ht]
    \centering
    \begin{tabular}{@{}llllll@{}}
    \toprule
        Dataset & $\varepsilon$  & Architecture& N Params& Validation Acc.  \\ \midrule 
        \multirow{5}{*}{CIFAR-10} & \multirow{5}{*}{$7.0$} & ResNet-18 & $11.2M$ & $59.8\%$\\
        && ResNet-34 & $21.2M$ & $57.7\%$ \\
        && DenseNet-121 & $7.0M$ & $65.4\%$\\
        && EfficientNet-B0 & $4.0M$ & $54.1\%$\\
        && SmoothNet (ours) &$3.4M$ & $\mathbf{73.5\%}$\\ \midrule
        \multirow{5}{*}{ImageNette} & \multirow{5}{*}{$7.0$} & ResNet-18 & $11.2M$ & $59.9\%$\\
        && ResNet-34 & $21.2M$ & $58.7\%$\\
        && DenseNet-121 & $7.0M$ & $65.1\%$ \\
        && EfficientNet-B0 & $4.0M$ & $60.5\%$\\
        && SmoothNet (ours) &  $3.4M$ & $\mathbf{69.7}\%$ \\
        \bottomrule
    \end{tabular}
    \caption{Benchmarking SmoothNet against standard architectures on CIFAR-10 and ImageNette with Differential Privacy. N Params refers to the number of trainable parameters in the network in Millions. Validation Acc. refers to the corresponding row's best model's accuracy on the validation set.}
    \label{tab:results}
\end{table*}
To validate our findings, we evaluated our novel architecture on two benchmark datasets. Firstly, CIFAR-10 served as a basis for evaluations with a long line of previous works. Moreover, to assess the performance of our architecture on a larger-image dataset, we also used ImageNette \cite{imagenette}, a subset of ImageNet, which --like the original dataset-- has a resolution of $224 \times 224$. We compared our architecture to several standard architectures, more specifically a small and medium sized ResNet (18, 34), a DenseNet-121. These architectures are the original architectures employing residual and dense connections respectively.  Lastly, we compared against EfficientNet-B0 \cite{tan2019efficientnet}, a high performance model with high parameter-efficiency. Similarly to the evaluation of specific model components, we trained each competing model 15 times with varying $L_2$-clipping norms in the range of $[0.1, 20.0]$ and for either 90, 120, or 180 epochs. We use Xavier-Glorot weight initialization \cite{Glorot2010UnderstandingTD}, a learning-rate decay of $0.9$ every five epochs, an initial-learning rate of $0.002$, an SGD momentum of $0.9$, a weight-decay of $0.0002$, and a batch-size of 256.
On both datasets, we found SmoothNets to achieve the highest performance in terms of validation accuracy (Table \ref{tab:results}). 

\begin{figure}
    \centering
    \includegraphics[width=0.45\textwidth]{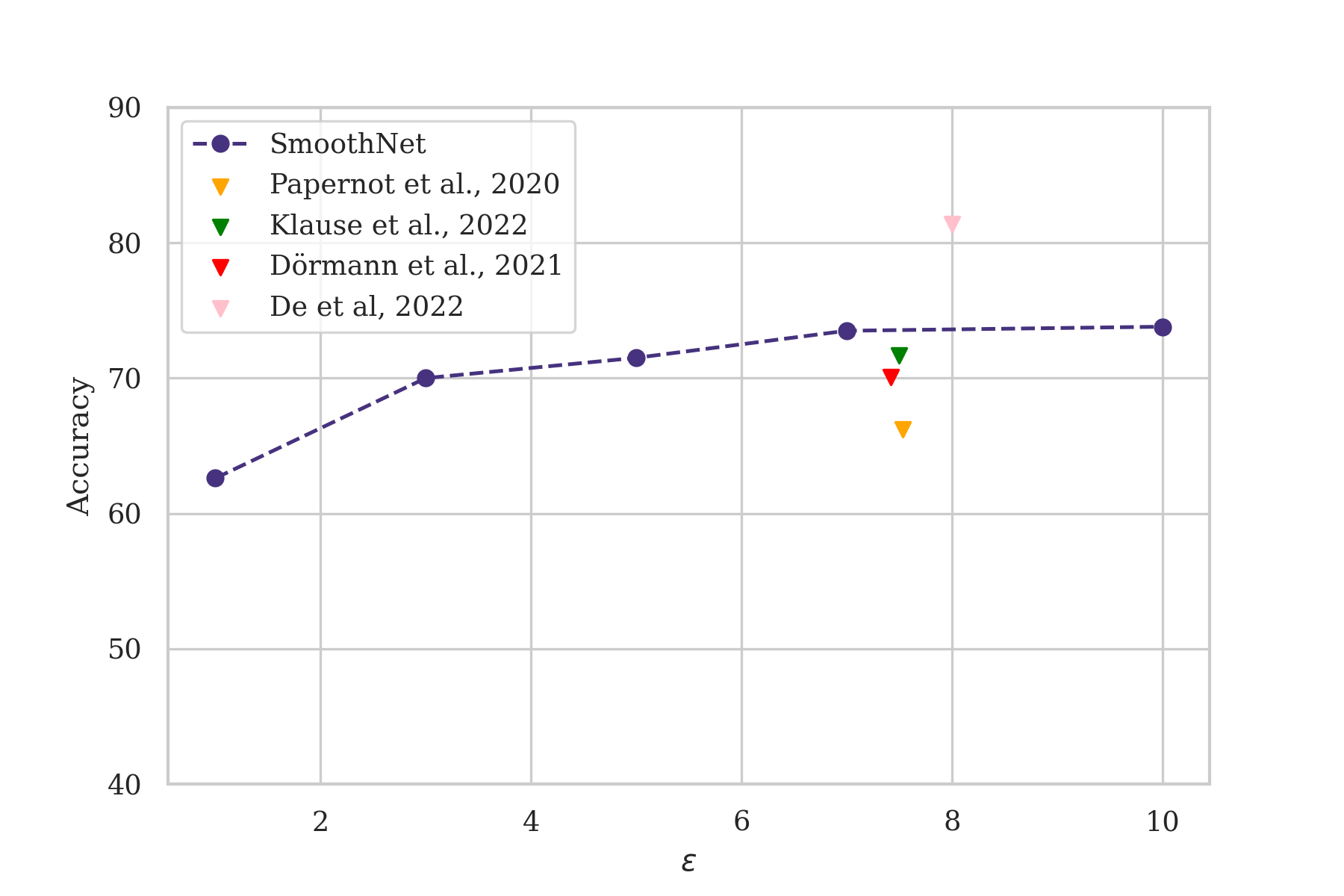}
    \caption{Pareto front contextualising our results compared to other work on the CIFAR-10 dataset.}
    \label{fig:pareto}
\end{figure}

\paragraph{Comparison to other works}
To contextualise our findings, we curated an overview of the results of related works in Table \ref{tab:other_works}. Furthermore, we investigate the Pareto front of our model over different values of $\varepsilon$ on CIFAR-10, which is visualised in Figure \ref{fig:pareto}, alongside results of other works. Our results outperform all previous works except \citealt{de2022unlocking}, which was released during the preparation of our manuscript. However, the authors of this work found that the greatest increase in accuracy resulted from an adaptation of the model's training instead of the architecture, hence we anticipate that our architectural modifications would act synergistically with the work of \citealt{de2022unlocking} to yield even higher performance, which we intend to evaluate in upcoming work. Interestingly, the authors of this work also employ WideResNets as an architecture, empirically corroborating our findings that wide models with residual connections improve training stability under \acs{DP} conditions. Of note, these results also motivate a more differentiated view on the claim that larger models are expected to have lower accuracy \textit{by default} because the DP-SGD noise scales proportionally to the number of parameters. Instead, model depth and width seem to interact in a more nuanced way with each other and with other architectural components.

\section{Discussion and Conclusion}
In this paper, we continue the investigation into optimal architectural choices for high-utility training of neural networks with DP guarantees. Prior works have demonstrated that specialised training methods and model adaptations can allow for high model utility under \acs{DP} guarantees. We naturally augment these previous results by introducing SmoothNet, an architecture that outperforms previous works focused on architectural modifications.

One limitation of our work is that, so far, we have no conclusive explanation for all of the contributing factors behind the high performance of our architecture in \acs{DPSGD}-training. We conjecture that the smoothness of the optimisation landscape is a strong influence on the performance. We observed that both dense connections and normalisation layers, which have been shown to improve this attribute \cite{li2018visualizing, santurkar2019does} to also improve model accuracy. This hypothesis also motivates the choice of name SmoothNet for our architecture. Moreover, the utilisation of SELU activation functions, stemming from previous work on normalisation-free architectures \citep{klambauer2017self}, seems to indicate that activation scale matters especially in DP-SGD. Similar findings are discussed in previous works \citep{papernot2020tempered, klause2022differentially}. We intend to empirically and theoretically investigate the training dynamics of SmoothNets in future work. Moreover, as our architectures allow for higher utility while preserving privacy budget, we intend to investigate their behaviour in the very-low $\varepsilon$ regime, which is --so far-- under-investigated in the realm of private deep learning.
In conclusion, our work is an empirical investigation into the factors enabling neural network training on sensitive datasets with high utility and strong privacy guarantees. We hope that our work will stimulate further research in this direction and elucidate the underlying reasons why specific model components are better suited for \acs{DP} training, which may lead to findings of independent theoretical interest.

\newpage
\bibliography{bibliography}
\bibliographystyle{icml2022}

\newpage
\appendix
\onecolumn


\end{document}